\documentclass{article}

\usepackage[preprint]{neurips_2025}

\usepackage[utf8]{inputenc}
\usepackage[T1]{fontenc}    
\usepackage{hyperref}       
\usepackage{url}            
\usepackage{booktabs}       
\usepackage{amsfonts}       
\usepackage{nicefrac}       
\usepackage{microtype}      
\usepackage{xcolor}         
\usepackage{graphicx}
\usepackage{subcaption}
\usepackage{float}
\usepackage{amsmath}

\title{Fusion Steering: Prompt-Specific Activation Control}

\author{
  Waldemar Chang \\
  Johns Hopkins University \\
  \texttt{wchang58@jh.edu} \\
  \And
  Dr. Alhassan Yasin \\
  Johns Hopkins University \\
  \texttt{ayasin1@jhu.edu}
}

\begin{document}

\maketitle

\begin{abstract}
We present \textit{Fusion Steering}, an activation steering methodology that improves factual accuracy in large language models (LLMs) for question-answering (QA) tasks. This approach introduces flexible steering configurations, including \textit{full-layer steering} and \textit{segmented steering}. Unlike traditional methods constrained to single-layer or fixed-layer operations, Fusion Steering employs dynamic injection of prompt-specific activation deltas across all transformer layers. These activation deltas are derived from reference completions that combine the ground-truth answer with a model-generated explanation to facilitate semantically enriched, example-specific steering. The injection weights are optimized per prompt using Optuna, targeting a joint objective that balances token overlap (factual alignment) and perplexity (fluency proxy). Evaluation employs a composite score integrating token overlap and LLM-graded quality, encompassing factual accuracy, coherence, and relevance. Empirical results on 260 SimpleQA prompts (selected from 500 where the baseline failed) showcase the efficacy of \textit{segmented steering}. Using Gemma-2-2B-IT with 8-bit quantization, \textit{segmented steering} achieves an accuracy of 25.4\% (outputs scoring $\geq 0.6$), outperforming the baseline at 3.5\% and \textit{full-layer steering} at 16.2\%. Under the stricter SimpleQA rubric, \textit{segmented steering} boosts fully correct responses from 0.0\% to 13.1\%. These findings highlight the strengths of segmented, dynamic intervention strategies and the promise of per-prompt, full-network activation control. Fusion Steering is also amenable to sparse representations, such as Neuronpedia or sparse crosscoders, suggesting a promising direction for interpretable and scalable activation-level control in LLMs.
\end{abstract}

\section{Introduction}

Large language models (LLMs) have exhibited exceptional capabilities across a broad spectrum of natural language processing tasks, spanning from fundamental text classification to intricate reasoning and question-answering (QA) \cite{brown2020languagemodelsfewshotlearners, openai2024gpt4technicalreport, touvron2023llamaopenefficientfoundation}. Nevertheless, achieving precise control over their outputs remains a formidable challenge. Recent advances in interpretability research underscore the complexity of comprehending or modulating LLM behavior at a holistic level, particularly due to long-range reasoning dependencies, abstract conceptual structures, and intertwined feature representations~\cite{lindsey2025biology}. This highlights the increasing importance of localized, prompt-specific control methodologies that integrate multi-layer interventions to mitigate cross-layer superposition and polysemantic feature entanglement~\cite{scaling_monosemanticity2024, crosscoders2024}. Prior investigations into steering neural network activations have predominantly centered on tasks involving straightforward semantic or syntactic adjustments, such as antonym generation, sentiment modulation, or toxicity suppression, where steering vectors are typically well-defined and exhibit linear characteristics within activation spaces~\cite{turner2024steeringlanguagemodelsactivation, postmus2025steeringlargelanguagemodels, panickssery2024steeringllama2contrastive}. Although effective in simpler contexts, these steering methods remain largely unexplored for more sophisticated, semantically rich tasks like factual question answering, which necessitate highly precise and context-aware outputs. All code and materials are available at \url{https://github.com/MechInterpreter/Fusion_Steering}.

In this work, we introduce a novel steering approach specifically designed for factual QA tasks. \textbf{Our method distinguishes itself from previous approaches in four key aspects:} 

\textbf{First}, we introduce the concept of \textit{enriched reference activations}, which capture neural activation vectors when the model processes a fully enriched prompt explicitly containing both the correct ground-truth answer and its explanation. These vectors encode detailed semantic information directly tied to factual accuracy and explanatory coherence. This approach differs from prior methods that derive latent steering vectors from generic prompts or contrastive pairs, without integrating explanatory context as a conditioning input~\cite{turner2024steeringlanguagemodelsactivation, panickssery2024steeringllama2contrastive, subramani2022extractinglatentsteeringvectors, konen2024stylevectorssteeringgenerative, stoehr2024activationscalingsteeringinterpreting}. 

\textbf{Second}, our approach employs a synergistic combination of interpolated activation patching and additive steering, executed concurrently across \textit{all} transformer layers. This comprehensive multilayer intervention strategy contrasts with prior methods, which typically applied steering to individual layers in isolation~\cite{turner2024steeringlanguagemodelsactivation, panickssery2024steeringllama2contrastive, heimersheim2024useinterpretactivationpatching, zou2025representationengineeringtopdownapproach}, rather than concurrently across the whole network. Furthermore, unlike previous methods, we integrate an interpolated variant of activation patching with the directional modulation introduced by additive steering, ensuring a more refined and context-sensitive control mechanism. 

\textbf{Third}, we introduce prompt- and layer-specific tuning by optimizing two hyperparameters per instance: the \textit{fusion weight}, which integrates the model’s original activations with the reference activations, and the \textit{steering strength}, which scales the magnitude of the injected activation shift. These parameters are independently tuned across three distinct layer groups—\textit{early} (layers 0–8), \textit{middle} (layers 9–17), and \textit{late} (layers 18–25), using 0-based indexing—enabling fine-grained, context-sensitive control over how factual guidance is incorporated throughout the model. Previous approaches to activation steering generally rely on a fixed intervention vector at a single layer or across multiple layers, often employing globally tuned intensity parameters that do not adapt to specific prompts or network regions~\cite{turner2024steeringlanguagemodelsactivation, panickssery2024steeringllama2contrastive, li2024inferencetimeinterventionelicitingtruthful, soo2025interpretablesteeringlargelanguage}. In contrast, our approach performs localized, per-prompt calibration at multiple functional depths of the model, offering more precise control over the factual alignment process. 

\textbf{Fourth}, while previous steering approaches have targeted specific behaviors or attributes with clearly defined latent directions~\cite{turner2024steeringlanguagemodelsactivation, postmus2025steeringlargelanguagemodels, panickssery2024steeringllama2contrastive, marshall2025refusalllmsaffinefunction}, our method is explicitly designed to enhance factual accuracy in open-ended question answering scenarios, where alignment cannot be captured by a single contrastive axis. This presents a more challenging control objective: the semantics of factual questions disperse across multiple, input-dependent dimensions. Effective control therefore requires semantically grounded steering capable of guiding the model toward discrete, knowledge-intensive answers. By distributing semantically meaningful reference activations throughout the network and dynamically adapting both their intensity and integration per prompt and layer group, our method enables holistic and precise control, explicitly guiding the model toward accurate, contextually grounded answers from the question alone.

We empirically validate our approach on the SimpleQA benchmark~\cite{wei2024measuringshortformfactualitylarge}, demonstrating that our method significantly improves factual accuracy compared to the base model’s responses.

\section{Methodology and experimental setup}

\textbf{Note:} All model generations in this work, whether from Gemma-2-2B-IT during inference or either Mistral Small 3.1 instance used for filtering and grading are performed under greedy decoding with \texttt{temperature = 0.0}. This ensures complete determinism throughout the pipeline and facilitates precise, reproducible evaluation across all experimental stages.

\subsection{Initial inference and subset construction}
We first ran the instruction-tuned baseline model (Gemma-2-2B-IT)~\cite{gemmateam2024gemma2} on 500 prompts from the SimpleQA test split and used a rubric-based factuality classifier built on the Mistral Small 3.1 model~\cite{mistral2024small} to grade each response as \texttt{CORRECT}, \texttt{INCORRECT}, or \texttt{NOT\_ATTEMPTED} (see Appendix~\ref{appendix:grading}). From these, we selected the 260 examples labeled \texttt{INCORRECT}, excluding both correct and non-attempted responses, for steering experimentation. We focus on this 260-prompt subset both due to computational constraints and because it represents the most challenging cases—each labeled \texttt{INCORRECT} by the LLM grader, making it ideal for stress-testing the effectiveness of steering interventions.

\begin{figure}[H]
    \centering
    \includegraphics[width=0.49\textwidth]{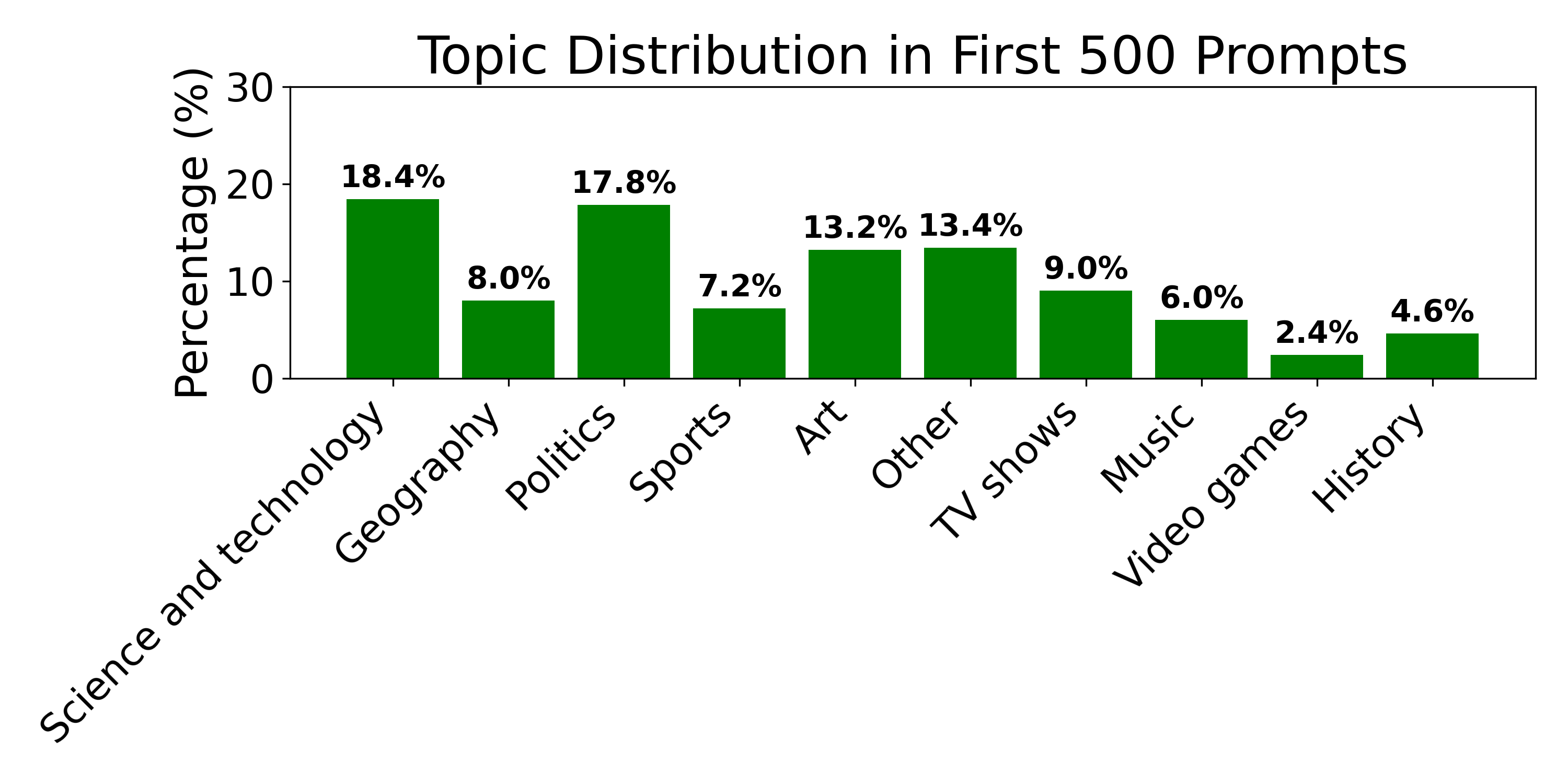}
    \includegraphics[width=0.49\textwidth]{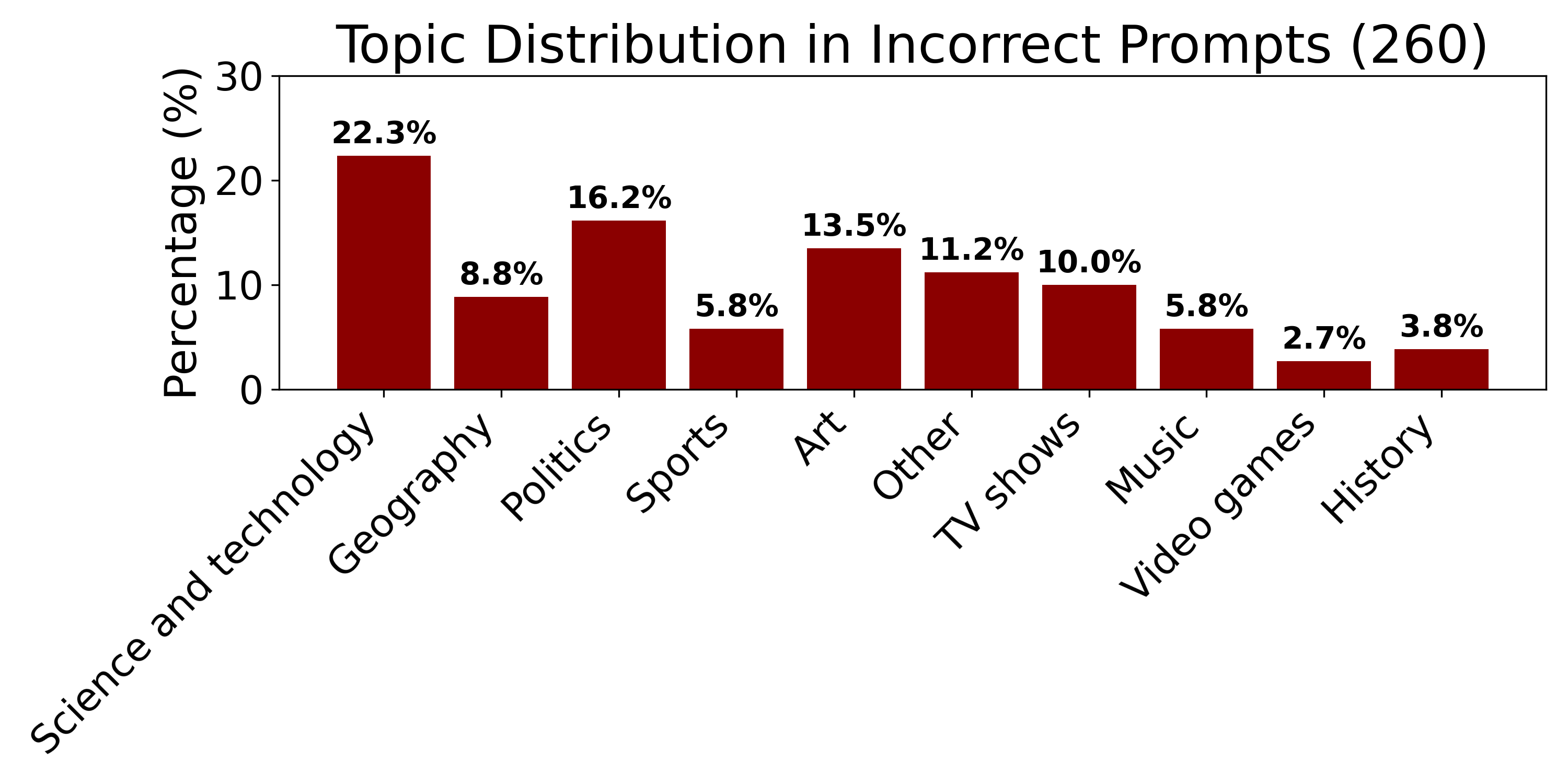}
    \includegraphics[width=0.49\textwidth]{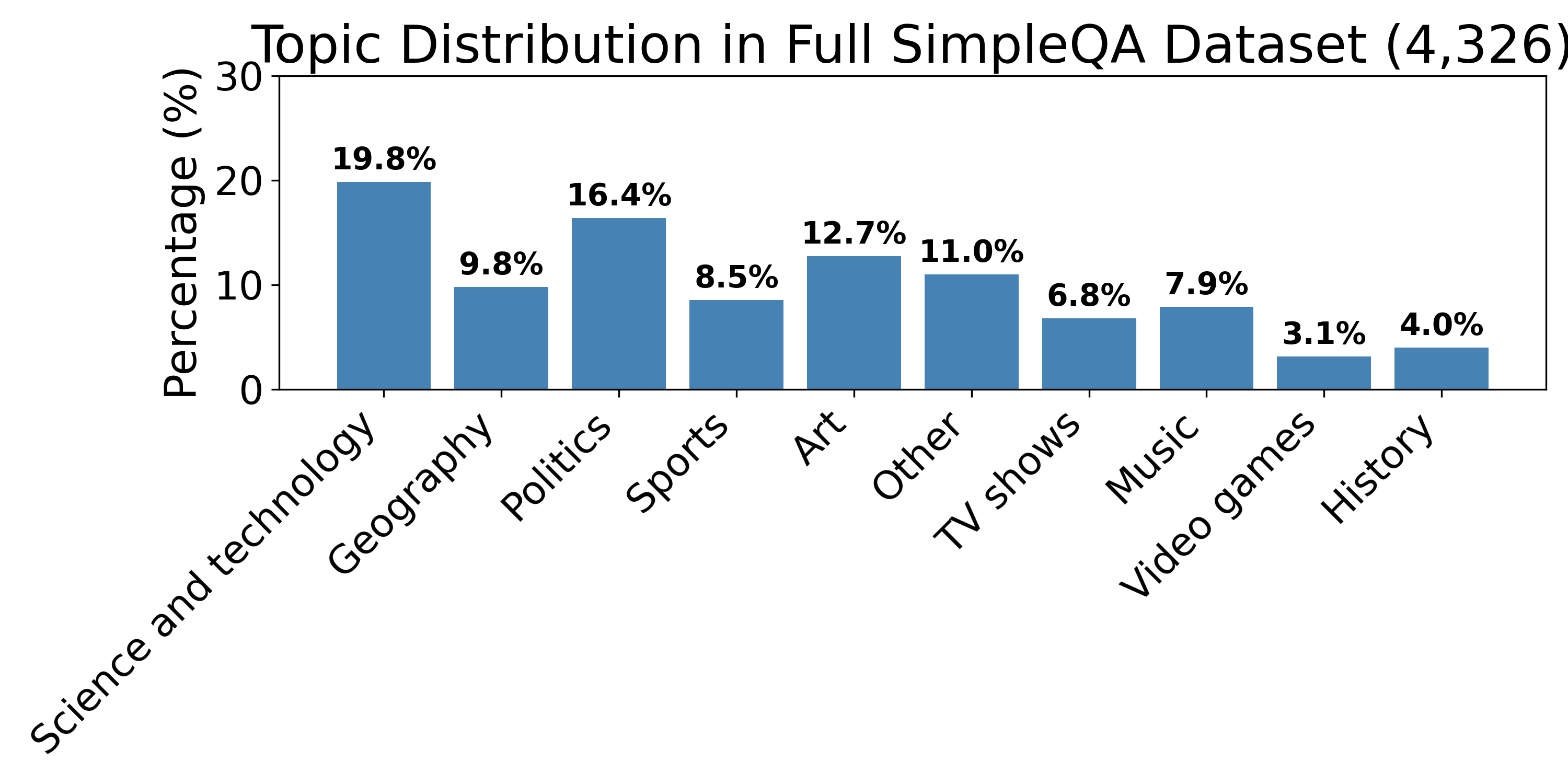}
    \caption{Topic distribution comparison across all subsets. Proportions remain consistent despite sample size differences, indicating strong stratification.}
    \label{fig:topic-distribution-panel}
\end{figure}

To confirm that our experimentation subset covers a representative range of topics, we compared the topic distribution of the full 500-prompt sample to the 260 \texttt{INCORRECT} examples. As shown in Figure~\ref{fig:topic-distribution-panel}, the distributions are broadly aligned across major categories such as \textbf{Science and Technology}, \textbf{Politics}, and \textbf{Art}, indicating that the incorrect subset retains the diversity of the larger sample.

Furthermore, the complete SimpleQA dataset spans 4,326 prompts in 10 topic categories, with the largest shares belonging to \textbf{Science and Technology} (19.8\%), \textbf{Politics} (16.4\%), and \textbf{Art} (12.7\%). This general distribution, shown in Figure~\ref{fig:topic-distribution-panel} and documented in the official release~\cite{wei2024measuringshortformfactualitylarge},\footnote{\url{https://openai.com/index/introducing-simpleqa/}} validates that our 500-sample—and by extension the 260-prompt subset—is well-stratified and representative of the broader benchmark.

\subsection{Capturing reference activations}

We capture reference activations from Gemma-2-2B-IT using an enriched prompt that combines the question, the ground-truth answer, and an explanation generated by the model after being fed the question and the ground truth:

\begin{quote}
\textbf{Question:} [prompt]\\
\textbf{Answer and Explanation:} [ground truth + explanation]
\end{quote}

This entire input sequence is tokenized and fed to the model. We then extract neural activations across all transformer layers, but only from token positions corresponding to the ground-truth answer and explanation. These activations are averaged at each layer to produce a single robust reference activation vector per layer. Formally, let \( h_l \in \mathbb{R}^d \) denote the activation vector for layer \( l \), calculated as:

{\small
\begin{equation}
h_l = \frac{1}{|\mathcal{T}|}\sum_{t \in \mathcal{T}} a_{l,t}
\end{equation}
}

where \( a_{l,t} \) denotes the activation at layer \( l \) for the token position \( t \), and \( \mathcal{T} \) denotes the set of token positions aligned to the ground-truth answer and explanation within the prompt.

\subsection{Multi-layer patching and steering}

At inference time, when presented only with the question (excluding the ground truth answer and explanation), we inject the previously captured reference activations across transformer layers using an additive steering mechanism. This steering dynamically nudges the internal representations toward the mean representation seen during reference activation capture.

Formally, let \( h_l \in \mathbb{R}^d \) denote the reference activation vector at layer \( l \), and let \( \mu_l \in \mathbb{R}^d \) be the mean of the activations at layer \( l \) across all token positions during inference. We compute a global shift vector:

{\small
\begin{equation}
\Delta_l = \gamma (h_l - \mu_l)
\end{equation}
}

This shift is added uniformly to each token's activation in the layer:

{\small
\begin{equation}
a_{l,t}^{\text{steered}} = a_{l,t}^{\text{original}} + \Delta_l
\end{equation}
}

Finally, we blend the steered and original activations using an interpolation parameter \( \alpha \in [0, 1] \), which we refer to as the \textit{fusion weight}:

{\small
\begin{equation}
a_{l,t}^{\text{final}} = \alpha a_{l,t}^{\text{steered}} + (1 - \alpha) a_{l,t}^{\text{original}}
\end{equation}
}

Here, \( \gamma \) modulates the intensity of the directional shift, and \( \alpha \) controls the extent of blending—the fusion step in our Fusion Steering method—between steered and original activations.

We explore two distinct implementations of this mechanism. In \textit{full-layer steering}, reference activations are injected uniformly across all transformer layers using a single set of $\alpha$ and $\gamma$ parameters. In contrast, \textit{segmented steering} divides the model into early, middle, and late layer groups, each steered using separately optimized $\alpha$ and $\gamma$ values. This allows for finer-grained control over how different representational stages of the model are influenced. We aim to understand the trade-offs between globally uniform steering and more targeted, segment-aware modulation of internal activations.

\subsection{Prompt-specific parameter optimization}

To ensure optimal parameter selection tailored to each input question, we employ Optuna-based hyperparameter optimization~\cite{akiba2019optunanextgenerationhyperparameteroptimization}. For each prompt-steering configuration, we ran five Optuna trials and selected the configuration with the highest composite score. Parameters are tuned separately for each prompt using a composite objective that captures both factual accuracy and linguistic fluency:

{\small
\begin{equation}
\text{Optimization Metric} = \lambda \cdot \text{Token Overlap} - (1 - \lambda) \cdot \text{Perplexity Normed}
\end{equation}
}

Here, \(\lambda \in [0, 1]\) governs the trade-off between factual alignment and fluency. The token overlap score captures n-gram-level agreement with the reference answer, while perplexity—measured via GPT-2~\cite{radford2019gpt2}—estimates generation fluency. Since lower perplexity indicates higher fluency, its normalized value is subtracted so that maximizing the objective favors both alignment and coherence.

We fix \(\lambda = 0.7\) to moderately prioritize factual alignment over fluency. While we did not perform an exhaustive sweep of \(\lambda\), this setting reflects a deliberate tradeoff: token overlap serves as a reliable optimization signal because SimpleQA ground-truths consist primarily of concise factual phrases. This setting was chosen heuristically as a stable midpoint between full token overlap weighting (\(\lambda = 1.0\)) and equal balance (\(\lambda = 0.5\)), based on pilot trials. More importantly, this design enables efficient per-prompt tuning, as both token overlap and perplexity can be computed rapidly. During optimization, the search space for both the fusion weight \( \alpha \) and steering strength \( \gamma \) is bounded to \([0.01, 1.0]\), ensuring sufficient expressiveness while avoiding degenerate extremes, and was found to offer stable convergence during pilot experiments.

In the \textit{full-layer steering} setup, the same \(\alpha\) and \(\gamma\) values are applied uniformly across all transformer layers for a given prompt. In contrast, \textit{segmented steering} partitions the model into early, middle, and late layer groups—defined as layers 0–8, 9–17, and 18–25 respectively, using 0-based indexing—with each group steered using its own optimized parameters. This finer-grained design allows more precise control over different stages of the model’s internal representation. While we apply this partitioning to Gemma-2-2B-IT, our choice is informed by prior work on LLaMA-style architectures showing that adjacent layers often share functional roles and exhibit high intra-group activation similarity \cite{sun2025transformerlayerspainters}. These findings suggest that transformer layers naturally cluster into coherent groups, supporting layer-wise modulation at group granularity even across different decoder-only architectures.

\subsection{Deterministic inference via greedy decoding}

To ensure consistency and eliminate variance due to sampling, all generations whether baseline, full-layer steering, or segmented steering—are produced using greedy decoding. At each timestep, the model deterministically selects the highest-probability token by setting \texttt{temperature = 0.0}; other sampling parameters (e.g., \texttt{top-p}) are left at default values and have no effect under this setting.

This inference mode guarantees a single, reproducible output for each prompt-steering configuration pair, allowing for precise attribution of differences in factuality or fluency to the effects of steering rather than stochastic variation. It also simplifies hyperparameter optimization by ensuring stable and reproducible metrics, such as token overlap and perplexity, for evaluating candidate configurations.

While greedy decoding may reduce output diversity, our primary focus is on factual correctness and reliability, both of which benefit from deterministic evaluation. Future work may extend this setup to sample-based decoding to assess the robustness and adaptability of activation-level steering under uncertainty.

\subsection{Evaluation metrics}

\subsubsection{Composite metric (token overlap and 5-point LLM evaluation)}
To comprehensively quantify the effectiveness of our steering methods, we compute a combined evaluation score that averages token overlap with a rubric-based LLM grade:
{\small
\begin{equation}
\text{Evaluation Metric} = 0.5 \cdot \text{Token Overlap} + 0.5 \cdot \left( \frac{\text{LLM Grade}}{5} \right)
\end{equation}
}
The LLM grade (scaled 1–5) is generated by Mistral Small 3.1~\cite{mistral2024small}, using a scoring rubric that evaluates the correctness, completeness, and relevance of generated responses (Appendix~\ref{appendix:llmgrading}). A numerical 1-5 rubric is used instead of letter grades to enable direct normalization and smooth integration into our continuous evaluation metric. Token overlap provides a granular, n-gram-level alignment check against concise ground-truth answers, serving as a lightweight but precise correctness proxy. By combining these complementary perspectives, our evaluation metric captures both surface-level lexical fidelity and deeper semantic accuracy, enabling more graded analysis of factual alignment even when predictions fall short of exact correctness.

\subsubsection{SimpleQA rubric grading}
To measure improvement relative to the original failure cases, we also apply a stricter SimpleQA rubric (Appendix~\ref{appendix:grading}) using the initial instance of Mistral Small 3.1, which was used to identify the 260 \texttt{INCORRECT} baseline completions. Unlike the 1–5 rubric, which enables finer-grained differentiation in output quality, including partial correctness and relevance, the SimpleQA rubric enforces a categorical distinction between fully correct, incorrect, and unattempted answers. While the composite metric captures incremental improvements in factual alignment and response quality, the SimpleQA rubric serves as a high-precision filter for absolute correctness. This combination allows us to assess steering impact both in terms of subtle semantic gains and strict correctness, offering complementary perspectives on factual performance.

\section{Results}

\subsection{Quantitative evaluation}

We evaluate steering effectiveness on 260 prompts where the baseline model initially failed (260 \texttt{INCORRECT} completions). A response is considered accurate if it achieves a composite score $\geq 0.6$, balancing token overlap and LLM-graded quality based on a 5-point rubric (Appendix~\ref{appendix:llmgrading}). Segmented steering achieves 25.4\% accuracy (66/260), compared to 16.2\% for full-layer steering and just 3.5\% for the baseline, representing a 7.3$\times$ improvement for segmented steering over the baseline. Distribution analysis (Figure~\ref{fig:score-box}) shows that segmented steering yields higher median and variance, indicating more robust and widespread gains across prompts.

\begin{figure}[H]
    \centering

    \begin{subfigure}{0.49\textwidth}
        \includegraphics[width=\linewidth]{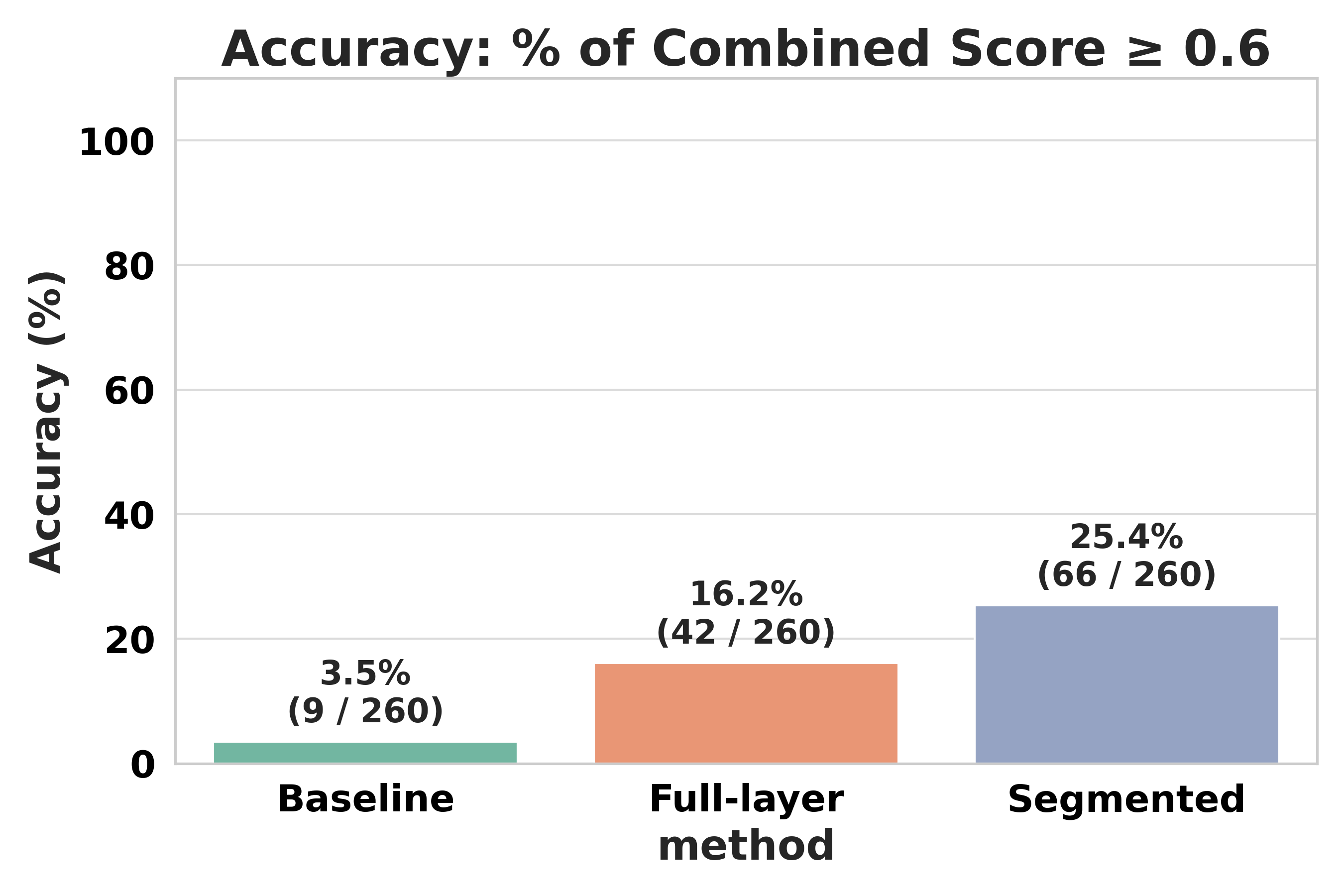}
        \caption{Combined Accuracy $\geq 0.6$}
        \label{fig:accuracy-bar}
    \end{subfigure}
    \hfill
    \begin{subfigure}{0.49\textwidth}
        \includegraphics[width=\linewidth]{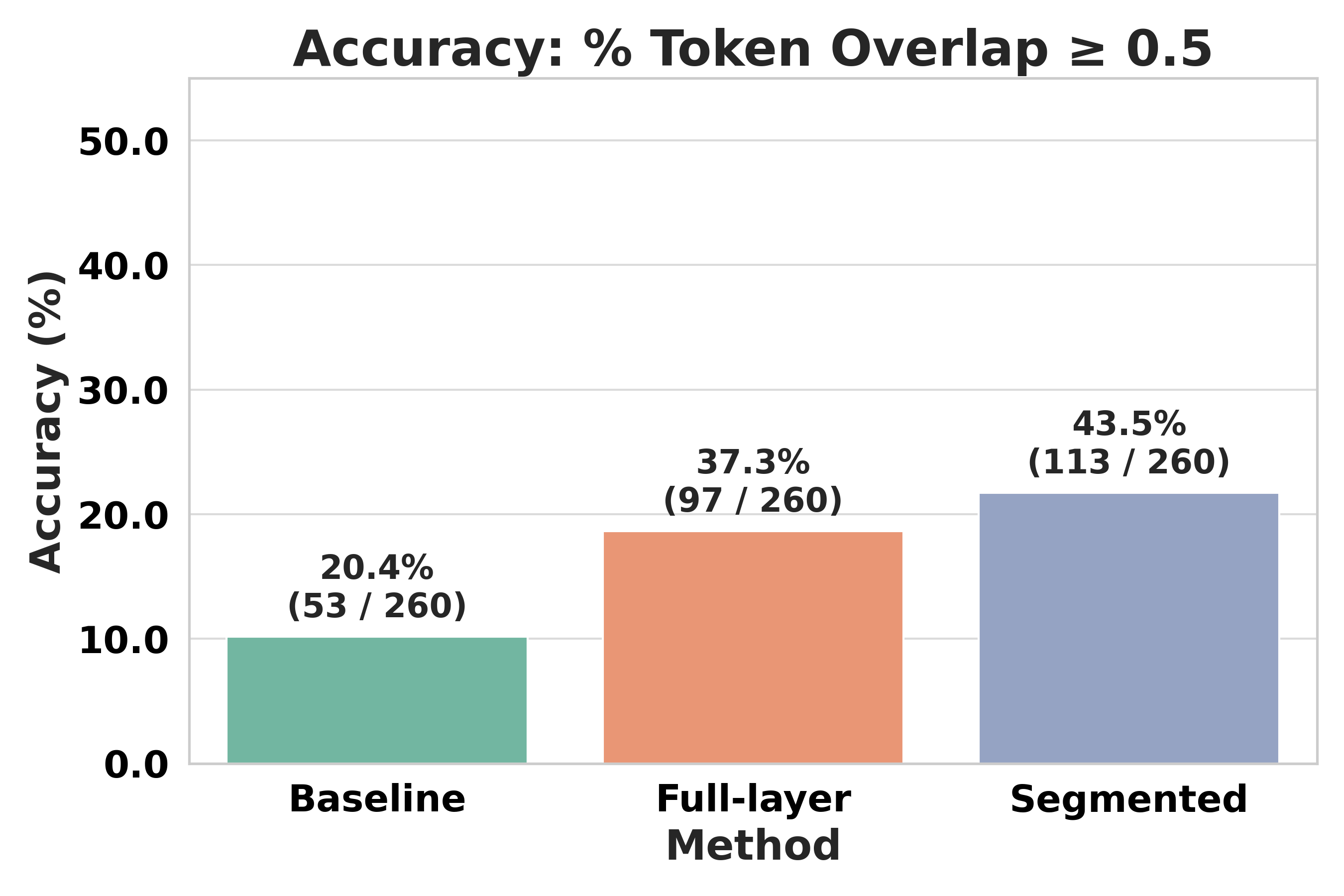}
        \caption{Token Overlap Accuracy $\geq 0.5$}
        \label{fig:token-overlap-bar}
    \end{subfigure}

    \vspace{0.5em}

    \begin{subfigure}{0.49\textwidth}
        \includegraphics[width=\linewidth]{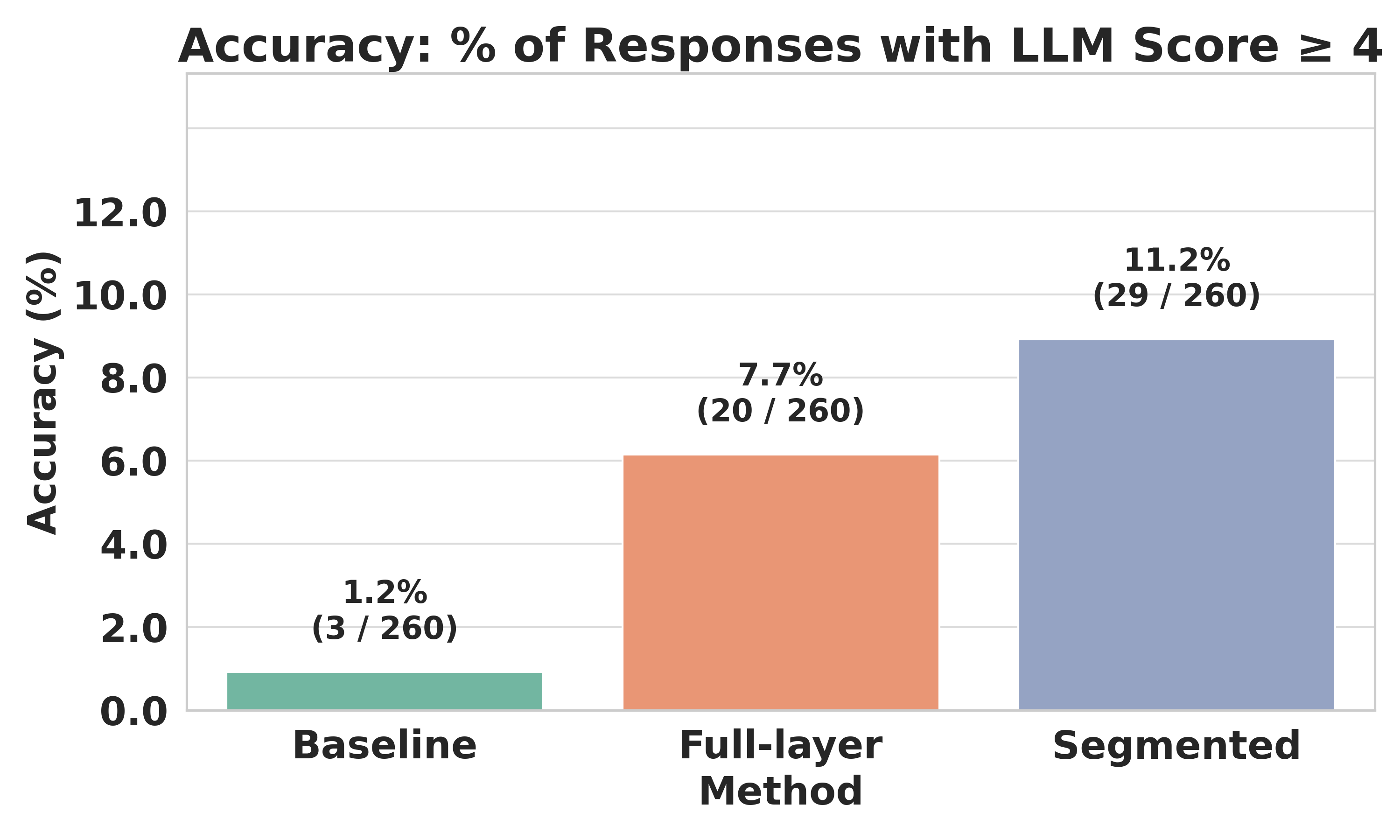}
        \caption{LLM Score Accuracy $\geq 4$}
        \label{fig:llm-acc-bar}
    \end{subfigure}
    \hfill
    \begin{subfigure}{0.49\textwidth}
        \includegraphics[width=\linewidth]{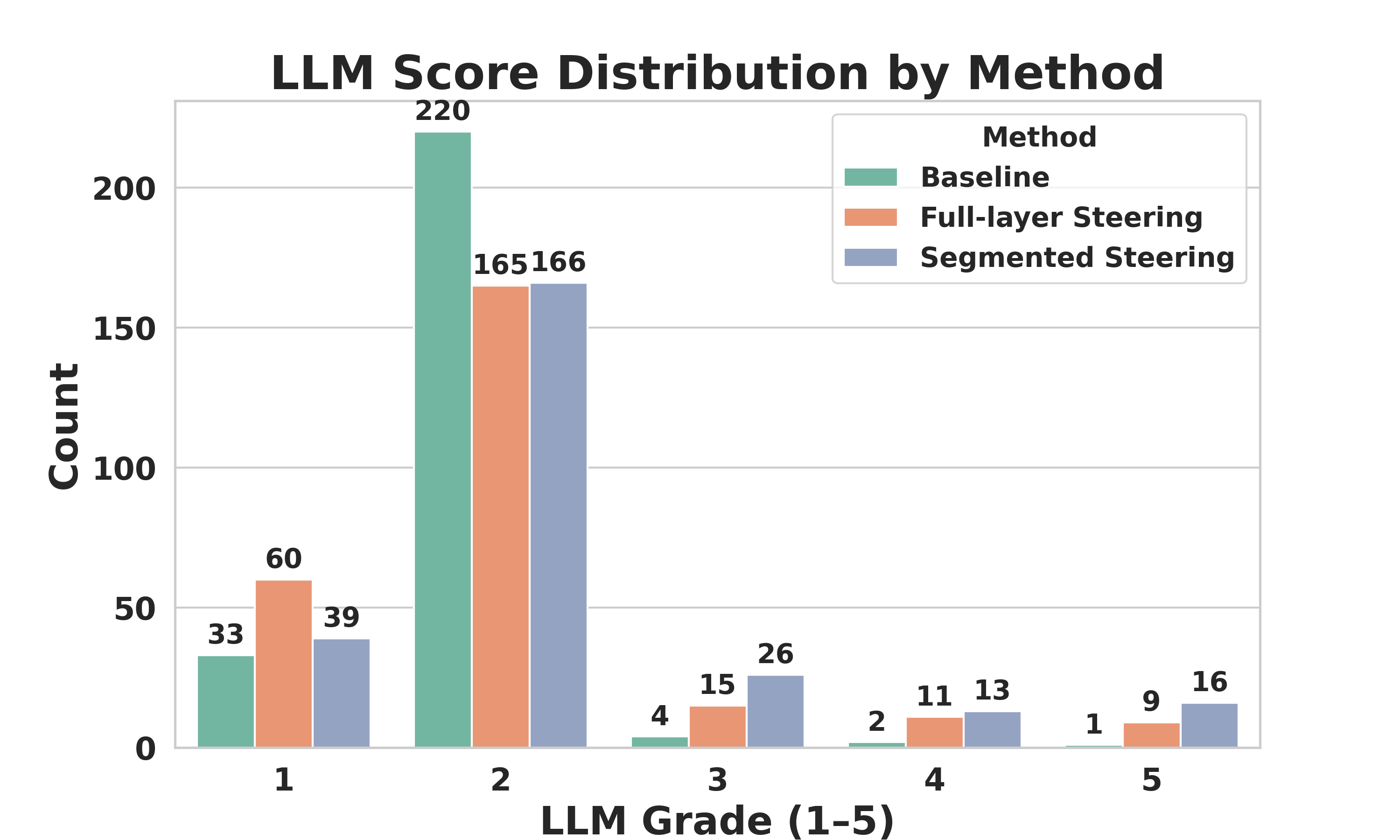}
        \caption{LLM Score Distribution (1–5)}
        \label{fig:llm-score-dist}
    \end{subfigure}

    \caption{
    Evaluation across three accuracy metrics. 
    (\textbf{Top Left}) Accuracy based on combined factual and quality score, 
    (\textbf{Top Right}) Accuracy based on token-level overlap, 
    (\textbf{Bottom Left}) Accuracy based on rubric-assigned LLM scores, and 
    (\textbf{Bottom Right}) Full distribution of LLM grades (1–5) assigned to model outputs. Segmented steering consistently outperforms all baselines across evaluation dimensions.
    }

    \label{fig:evaluation-core}
\end{figure}

For token overlap $\geq 0.5$, segmented steering again leads with 43.5\% accuracy (113/260), outperforming full-layer (37.3\%) and baseline (20.4\%). It also attains the highest mean (0.38), median (0.23), and variance (0.42) of overlap scores (Figure~\ref{fig:token-overlap-box}), highlighting its consistency across diverse QA topics. Notably, the SimpleQA benchmark spans a wide range of domains like science, politics, art, and history, suggesting that these gains generalize across semantically distinct question types.

We further validate factual correctness using a stricter SimpleQA LLM grading scheme (Appendix~\ref{appendix:grading}) that categorizes responses as \texttt{CORRECT}, \texttt{INCORRECT}, or \texttt{NOT\_ATTEMPTED}. Segmented steering achieves 13.1\% \texttt{CORRECT} (34/260), full-layer achieves 9.2\% (24/260), and the baseline produces zero \texttt{CORRECT} completions (Figure~\ref{fig:simpleqa-bar}). This discrete evaluation confirms that steering not only boosts composite scores but also enables fully correct generations under a more conservative rubric.

To unpack the more nuanced rubric-based LLM scores (Appendix~\ref{appendix:llmgrading}), we analyze the full distribution of 1–5 quality ratings (Figure~\ref{fig:llm-score-dist}). Surprisingly, both steering variants yield a higher number of completions graded 1 than the baseline. This reflects a shift in failure mode: while the baseline often generates fluent but factually irrelevant responses (typically graded 2), the steered variants push the model closer to the target answer, sometimes at the expense of fluency or syntactic coherence.

This dynamic illustrates a key trade-off in low-temperature decoding: steering increases alignment pressure, but without fluency-preserving mechanisms, may also degrade coherence. This is particularly evident in completions where the model's syntax breaks down as it attempts to emulate the ground-truth sequence (see Table~\ref{tab:circle-vince-comparison}).

Despite this, both full-layer and segmented steering significantly increase the number of completions rated 3, 4, or 5—bins the baseline almost never reaches. This redistribution highlights a qualitative shift: instead of defaulting to safe but incorrect outputs, the steered generations aim for factual correctness, raising the ceiling of answer quality even as they widen the distribution of outcomes.

\subsection{Qualitative analysis}

To better illustrate how Fusion Steering impacts factual accuracy and response quality, we provide representative examples comparing baseline outputs and steered generations in Table~\ref{tab:el-anatsui-comparison}. These samples highlight typical improvements, particularly in factual correctness and coherence. All prompt completions, including baseline, full-layer, and segmented outputs across all 260 examples, available for review in the supplementary CSV files, ensuring transparency and reproducibility.

The qualitative example in Table~\ref{tab:el-anatsui-comparison} illustrates several key dynamics enabled by Fusion Steering. The baseline model selects an incorrect answer (University of London) and provides a plausible but factually inaccurate justification. Notably, the baseline begins by hallucinating a multiple-choice format, listing four plausible universities before selecting one, despite the prompt not requesting such structure. None of the listed options include the correct answer (Harvard University), indicating that the model’s retrieval or grounding mechanisms fail to access the relevant fact at all. Full-layer steering attempts a correction, shifting to a different plausible university (MIT) and enriching the explanation with thematic relevance to El Anatsui’s artistic work. However, the selected institution remains incorrect, suggesting that full-layer activation shifts may overgeneralize or retrieve semantically adjacent but factually incorrect entities.

In contrast, segmented steering successfully grounds the generation in the correct answer (Harvard University) and retains coherence and justification quality. This supports the notion that segmented steering enables more precise factual targeting, likely due to its fine-grained control over the model’s internal representations. Although both steered outputs improve contextual grounding and explanation detail relative to the baseline, segmented steering demonstrates superior factual alignment.

\begin{table}[H]
\centering
\small
\renewcommand{\arraystretch}{1.2}
\caption{Comparison of baseline, full-layer, and segmented steering outputs for a factual question. All outputs are shown verbatim and truncated for space.}
\label{tab:el-anatsui-comparison}

\vspace{0.5em}

\begin{tabular}{p{13cm}}
\toprule
\textbf{Prompt 61} \hfill \textit{Ground Truth: Harvard University} \\
\midrule
Which university gave El Anatsui an honorary doctorate in 2016? \\
\bottomrule
\end{tabular}

\vspace{0.75em}

\begin{tabular}{p{4.2cm} p{4.2cm} p{4.2cm}}
\toprule
\textbf{Baseline Output} & \textbf{Full-layer Steering Output} & \textbf{Segmented Steering Output} \\
\midrule
A. University of Oxford  
B. University of Cambridge  
C. University of Edinburgh  
D. University of London  
\newline
\textbf{Answer:} D. University of London  
\newline
El Anatsui is a Ghanaian artist known for his large-scale installations made from discarded bottle caps. He received an honorary doctorate... &

El Anatsui, a renowned Ghanaian artist known for his innovative use of discarded materials, received an honorary doctorate from \textbf{Massachusetts Institute of Technology (MIT)} in 2016. His work, which often incorporates metal, textiles, and other found objects... &

El Anatsui, a renowned Ghanaian artist, received an honorary doctorate from \textbf{Harvard University} in 2016. His innovative use of recycled materials, particularly metal, in creating large-scale installations... \\
\bottomrule
\end{tabular}
\end{table}

This example illustrates how activation-level interventions—especially when guided by layer-specific hyperparameter tuning—can more precisely steer model behavior toward truth-consistent completions. By tailoring steering strength and integration depth across different transformer layers, the model is better able to align with factual targets. At the same time, the example reveals trade-offs between factual precision and semantic drift, highlighting the value of dynamic, context-sensitive control over internal representations.

The example shown below in Table~\ref{tab:circle-vince-comparison} (Prompt 38) highlights the nuanced interaction between steering-driven factual alignment and surface-level fluency. The baseline model and full-layer steering both provide fluent yet incorrect responses, confidently stating "Episode 4" rather than the ground-truth "Episode 7." By contrast, segmented steering attempts to align closely with the correct factual information, repeatedly generating variations of the correct token ("7"). However, this aggressive push toward the correct lexical content leads to a severe degradation in coherence, resulting in a response scored as the lowest grade (1) by the LLM rubric.

\begin{table}[H]
\centering
\small
\renewcommand{\arraystretch}{1.2}
\caption{Comparison of outputs for a factual recall question with ground-truth answer ``7.'' Baseline and full-layer responses are fluent but factually incorrect, while segmented steering approximates the correct answer but fails in surface coherence.}
\label{tab:circle-vince-comparison}

\vspace{0.5em}

\begin{tabular}{p{13cm}}
\toprule
\textbf{Prompt 38} \hfill \textit{Ground Truth: 7} \\
\midrule
In Season 3 of the American version of ``The Circle,'' in which episode did Vince enter the game? \\
\bottomrule
\end{tabular}

\vspace{0.75em}

\begin{tabular}{p{4.2cm} p{4.2cm} p{4.2cm}}
\toprule
\textbf{Baseline Output} & \textbf{Full-layer Steering Output} & \textbf{Segmented Steering Output} \\
\midrule
The answer is \textbf{Episode 4}. &

Vince, a popular player in Season 3, entered the game in \textbf{Episode 4}. Here's why: The Circle is known for its strategic gameplay and twists... &

............................7...7..7.7.7.7.7... \\
\bottomrule
\end{tabular}
\end{table}

This fluency breakdown may be partly attributable to the combination of greedy decoding and the nature of token-position expectations during activation extraction. Specifically, activations were extracted not only from ground-truth answer tokens but also from extensive explanation tokens, implicitly conditioning the model to fill in a correspondingly long explanation during generation. Coupled with greedy decoding and a relatively large token-generation budget (\texttt{max\_new\_tokens=100}), the model becomes locked into repetitively generating the token most strongly associated with correctness (``7''), despite a lack of coherent context.

This case vividly illustrates the seemingly counterintuitive pattern discussed earlier: segmented steering methods frequently yield higher token overlap with ground truth, indicative of improved factual alignment, yet may simultaneously receive lower rubric scores due to compromised fluency and readability. Such examples explain the widened distribution of LLM rubric scores observed in Figure~\ref{fig:llm-score-dist}, where segmented steering shifts outputs away from mid-range scores toward the extremes, achieving higher factual accuracy in some instances but at the cost of coherence in others. The phenomenon underscores the inherent trade-off steering methods navigate between factual correctness and surface-level quality.

\textbf{Note:} Statistical significance testing was not performed due to the deterministic nature of our evaluation setup. All completions were generated via greedy decoding, and steering parameters were tuned per prompt, eliminating stochastic variation across runs.

\section{Computational resources}

All experiments, including per-prompt hyperparameter tuning, activation extraction, and steering inference, were conducted on a single NVIDIA T4 GPU, representative of commonly available, resource-constrained environments. To reduce memory consumption and accelerate computation, we employed 8-bit quantization~\cite{nagel2021whitepaperneuralnetwork} via the HuggingFace \texttt{bitsandbytes} integration, enabling over 2,600 generations across tuning and evaluation phases without exceeding hardware limitations. To further ensure computational efficiency, all hyperparameter optimization was performed using Optuna~\cite{akiba2019optunanextgenerationhyperparameteroptimization}, leveraging efficient search algorithms to minimize optimization overhead.

This setup highlights the lightweight and scalable nature of Fusion Steering: our method operates entirely at inference time, requires no model fine-tuning or retraining, and performs effectively on modest hardware. Despite limited resources, we observe substantial gains in factual accuracy, demonstrating that internal activation-level intervention can be both efficient and impactful. The approach is fully reproducible, with open-source code and minimal system requirements, making it well-suited for adoption in resource-limited research contexts and rapid prototyping workflows.

\section{Limitations and future directions}
While Fusion Steering demonstrates promising improvements in factual accuracy, several limitations must be acknowledged. First, the method currently operates under greedy decoding (\texttt{temperature = 0.0}), which, while ensuring deterministic outputs for controlled evaluation, may exacerbate brittleness and fluency degradation in edge cases. Second, our steering signals are derived from reference activations that include both ground-truth answers and explanations; this conditioning may implicitly encourage overly long or unnatural completions when explanation tokens dominate the activation space.

However, in practical deployments, such enriched reference prompts are often unavailable or incomplete. The requirement of precise ground-truth answers and detailed explanations limits the scalability of Fusion Steering in low-resource or real-time environments where curated supervision is rare. This reliance also poses challenges for domains involving open-ended or multi-hop reasoning, where ground-truths may be ambiguous or unstructured.

Additionally, the method is evaluated only on a subset of 260 incorrect prompts from the SimpleQA benchmark, limiting generalizability to broader QA or open-domain generation settings. The optimization process is prompt-specific, requiring per-instance hyperparameter tuning, which poses computational challenges in high-throughput or streaming scenarios. We also do not explore alternate decoding strategies (e.g., beam search, sampling) or assess robustness to adversarial or noisy activations.

These limitations motivate future work on plug-and-play steering mechanisms that do not require full reference completions. One promising direction is to integrate Fusion Steering with emerging neuron-level semantic representations, such as those produced by sparse autoencoders (e.g., Neuronpedia) \cite{paulo2024automaticallyinterpretingmillionsfeatures} or crosscoder-based alignment methods \cite{crosscoders2024}, which identify high-level concepts within specific units. Rather than injecting raw activations derived from entire answer–explanation sequences, future systems could extract interpretable features directly from the input prompt, then activate neuron subsets that correspond to relevant semantic regions of activation space. If the geometry of concept space becomes reliably mapped, where discrete features are tied to identifiable neuron clusters, then prompts could be enriched by embedding structured encodings of these features alongside the input. This would shift steering from recall-based intervention to a feature-driven, concept-conditioned mechanism, enabling scalable and interpretable factual alignment even in open-domain or under-specified generation settings.

Recent work provides encouraging theoretical and empirical foundations for this direction. The “Geometry of Concepts” study \cite{Li_2025} demonstrates that sparse autoencoders produce structured representations, where related concepts cluster spatially in activation space, forming modular and hierarchical geometric patterns such as parallelograms and trapezoids. Similarly, Anthropic’s Scaling Monosemanticity project \cite{scaling_monosemanticity2024} shows that sparse autoencoders can extract interpretable features from large language models, with semantically similar concepts clustering in localized neuron groupings across modalities and languages. Sparse crosscoders \cite{crosscoders2024} extend this approach by analyzing features distributed across layers, revealing shared structures and potentially simplifying circuit analysis through cross-layer superposition. These advances suggest that future variants of Fusion Steering could selectively activate semantically-relevant feature neighborhoods instead of full reference traces, enabling more interpretable and concept-conditioned factual intervention, with potential trade-offs between transparency and computational overhead.

\section{Conclusion}
We introduced Fusion Steering, a novel activation-level intervention technique designed to improve factual accuracy in large language models for open-ended question answering. Our method combines enriched reference activations, multi-layer interpolated activation patching and additive steering, and prompt- and layer-specific hyperparameter tuning to guide model behavior at inference time. Unlike prior approaches targeting tasks with clearly defined linear directions, Fusion Steering is tailored for discrete, knowledge-intensive QA, where steering must guide the model toward specific factual entities rather than general stylistic shifts. It enables semantically grounded guidance without modifying model weights or requiring fine-tuning. Empirical results on 260 failed SimpleQA prompts show that segmented steering outperforms both baseline and full-layer methods, achieving over 7-fold accuracy gains under the composite metric, and an absolute improvement from 0.0\% to 13.1\% under the SimpleQA rubric. These results suggest that prompt-specific, interpretable activation interventions offer a promising direction for enhancing factual alignment in language models. Future work may explore how Fusion Steering integrates with neuron-level interpretability frameworks to enable even more targeted, concept-conditioned interventions.

\newpage
\bibliographystyle{unsrt}
\bibliography{references}

\newpage
\appendix
\section*{Appendix}
\addcontentsline{toc}{section}{Appendix}
\section{Factuality grading prompt template}
\label{appendix:grading}

We used a rubric-based prompt with the Mistral-small model to classify predicted answers as 

\texttt{CORRECT}, \texttt{INCORRECT}, or \texttt{NOT\_ATTEMPTED}. Below is the full grading prompt used for evaluation:

\begin{center}
\small
\begin{verbatim}
You are a factuality grader. Given a question, a gold target (correct answer), 
and a predicted answer, assign one of the following grades:
- "CORRECT" (fully contains correct information, no contradictions)
- "INCORRECT" (contains incorrect, misleading, or contradictory information)
- "NOT_ATTEMPTED" (fails to answer or provides irrelevant information)

### Example Grades:
CORRECT:
- Question: "Who discovered gravity?"
  - Gold target: "Isaac Newton"
  - Predicted: "Isaac Newton"
  - Predicted: "Newton discovered gravity."
  - Predicted: "Most believe Isaac Newton discovered gravity."

INCORRECT:
- Question: "Who discovered gravity?"
  - Gold target: "Isaac Newton"
  - Predicted: "Albert Einstein"
  - Predicted: "Newton, Einstein, and Galileo discovered gravity." 
  (Contradicts correct answer)
  - Predicted: "Gravity was first theorized in 1905." 
  (Incorrect)

NOT_ATTEMPTED:
- Question: "Who discovered gravity?"
  - Gold target: "Isaac Newton"
  - Predicted: "I don't know."
  - Predicted: "It depends on interpretation."
  - Predicted: "Without more context, I cannot say."

### Rules:
- Ignore capitalization, punctuation, and grammar differences.
- Allow hedging (e.g., "I think" or "probably") as long as the correct answer is included.
- If a predicted answer partially matches but is missing key details, grade as INCORRECT.
- If a number is the correct answer, it must be accurate to the last significant figure.
- Do not punish minor typos in names if the intent is clear.

### Now, grade this:
Question: {question}
Gold target: {target}
Predicted answer: {predicted_answer}

Reply with:
A: CORRECT
B: INCORRECT
C: NOT_ATTEMPTED

Only return "A", "B", or "C" with no extra text.
\end{verbatim}
\end{center}

\newpage
\section{LLM-based quality scoring rubric}
\label{appendix:llmgrading}

To evaluate the quality of model outputs after steering, we used a second rubric-driven prompt with the Mistral-small model. This rubric assigns a 1–5 score based on correctness, completeness, and relevance, and also records hallucination flags and short justifications. The full prompt used is shown below:

\begin{center}
\small
\begin{verbatim}
You are a strict but fair evaluator. 
Evaluate the generated answer for correctness, relevance, and completeness.

Question: {question}
Generated Answer: {generated_text}
Ground Truth Answer: {ground_truth}

Give a score from 1 to 5 based on the following:
- 5 = Fully correct, complete, and relevant.
- 4 = Mostly correct, with minor inaccuracies or missing details.
- 3 = Partially correct, has significant omissions or minor errors.
- 2 = Mostly incorrect or irrelevant, with small fragments that match.
- 1 = Totally wrong, irrelevant, or hallucinated.

Also specify:
- Hallucination (Y/N): Did the answer include fabricated information not found in the 
ground truth?
- Notes: Brief justification.
\end{verbatim}
\end{center}

\section{Score distribution plots}
\label{appendix:distribution}

\begin{figure}[H]
    \centering

    \begin{subfigure}{0.5\textwidth}
        \includegraphics[width=\linewidth]{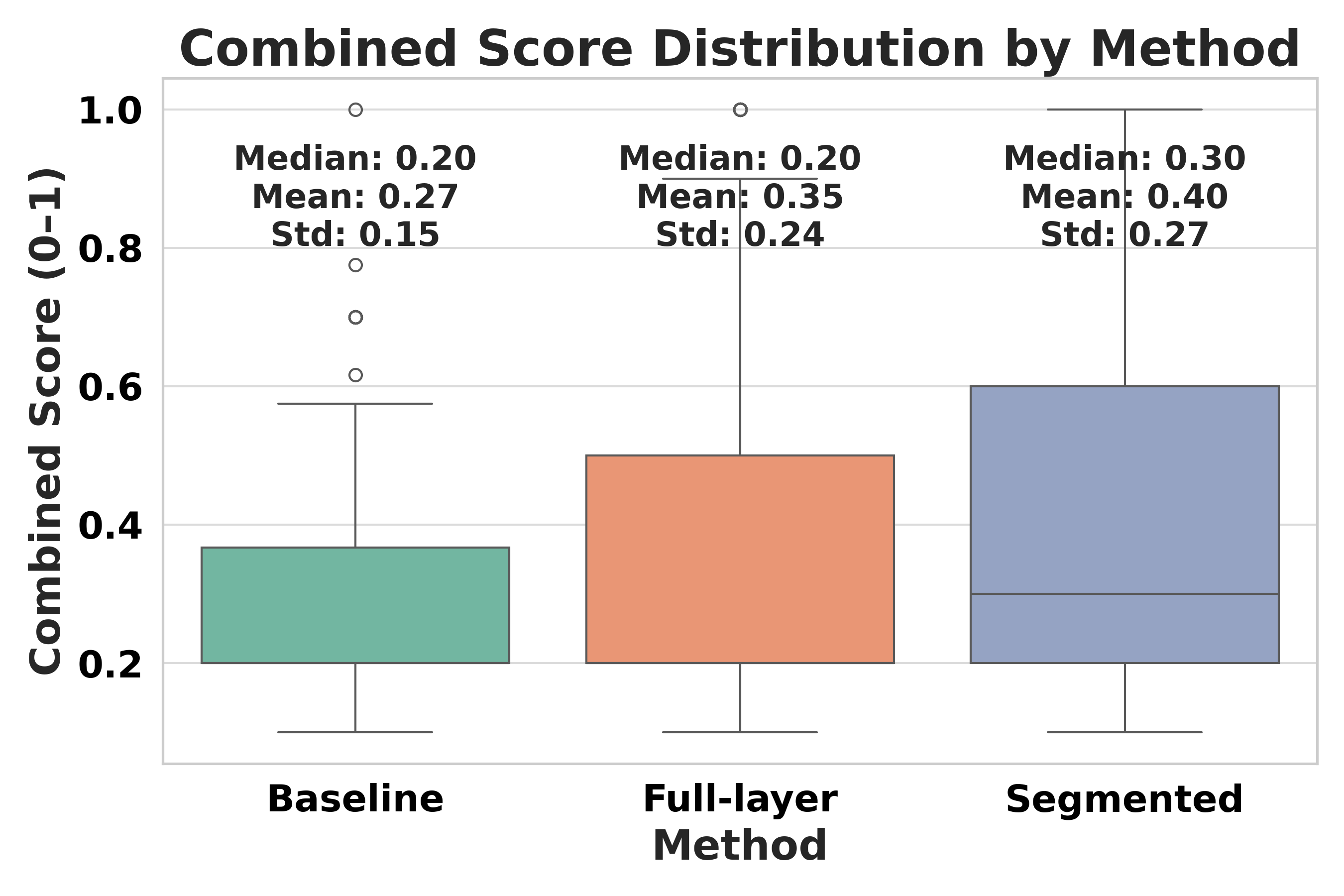}
        \caption{Combined Score Distribution}
        \label{fig:score-box}
    \end{subfigure}

    \vspace{1em}

    \begin{subfigure}{0.5\textwidth}
        \includegraphics[width=\linewidth]{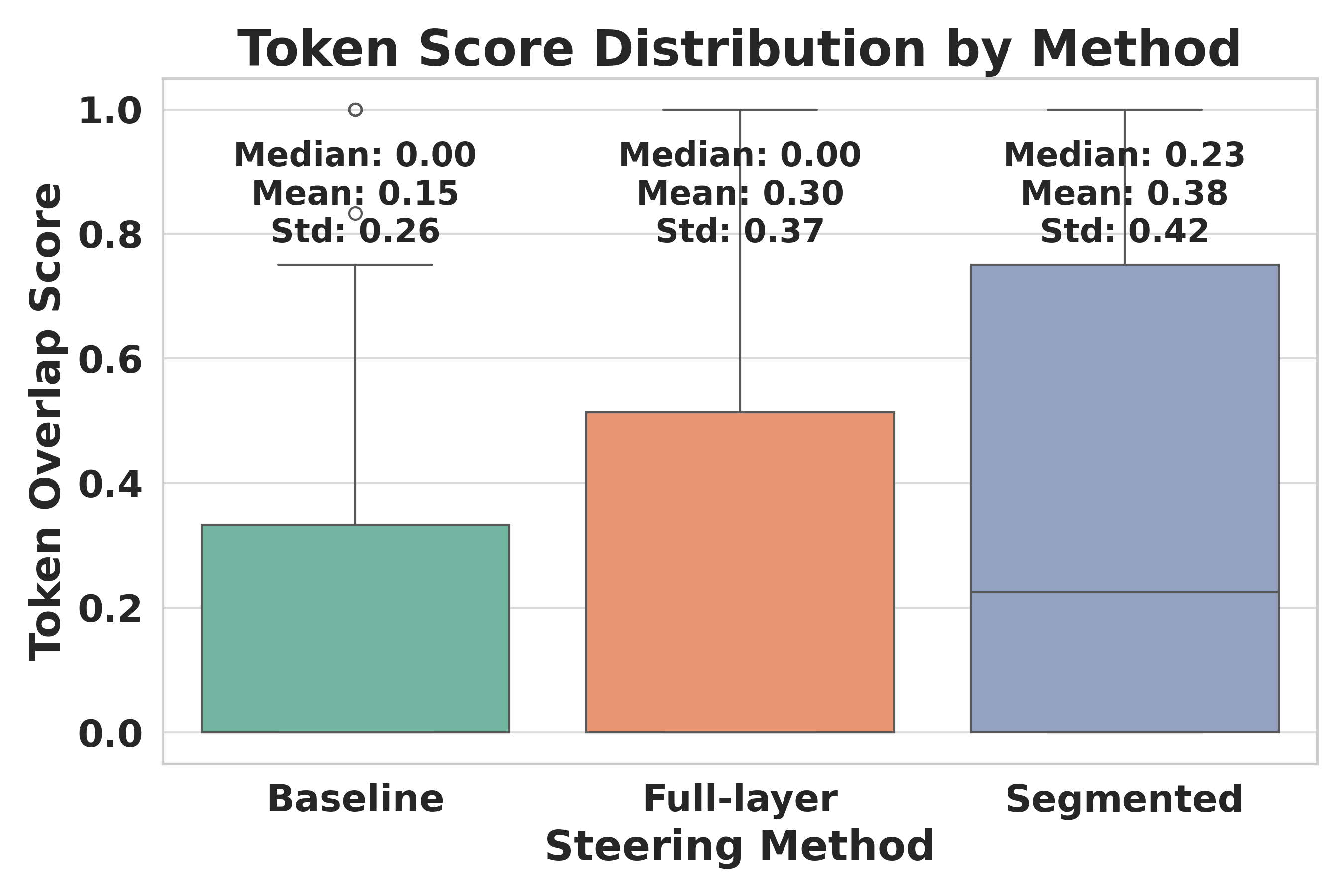}
        \caption{Token Overlap Score Distribution}
        \label{fig:token-overlap-box}
    \end{subfigure}

    \caption{
    Score distributions for combined metric and token overlap. 
    Boxplots illustrate variation in steering effectiveness across prompts.
    }
    \label{fig:appendix-score-distributions}
\end{figure}

\section{SimpleQA rubric grading accuracy}
\label{appendix:simpleqa-barplot}

\begin{figure}[H]
    \centering
    \includegraphics[width=0.65\linewidth]{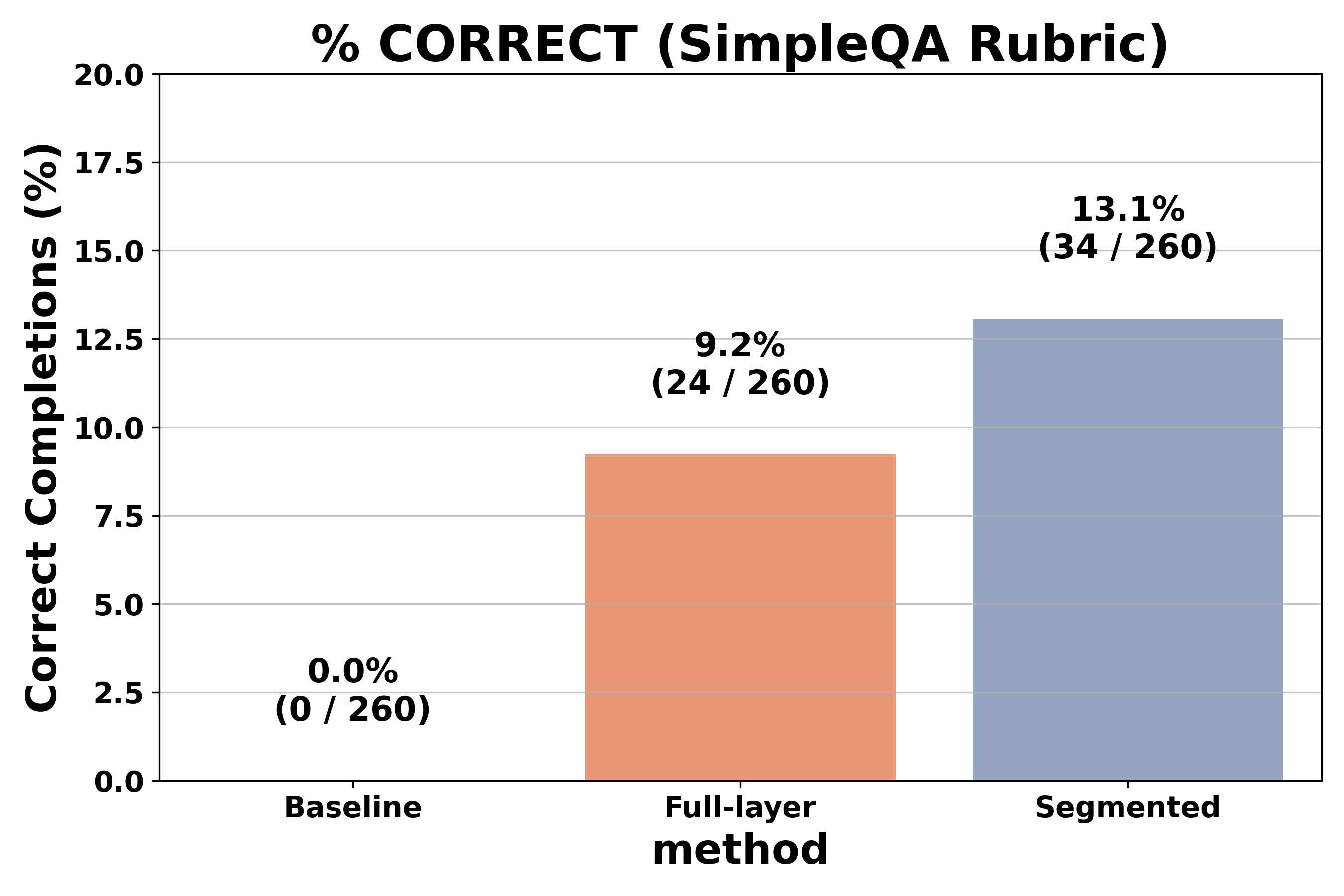}
    \caption{
    Percentage of completions labeled as \texttt{CORRECT} under the SimpleQA rubric grading. Bar heights represent the proportion of examples (out of 260) graded as fully correct by an LLM-based evaluator. The baseline yielded no correct responses, while full-layer and segmented steering show substantial gains.
    }
    \label{fig:simpleqa-bar}
\end{figure}

\end{document}